# Partitioned Linear Programming Approximations for MDPs

**Branislav Kveton**
Intel Research
Santa Clara, CA
*branislav.kveton@intel.com*

**Milos Hauskrecht**
Department of Computer Science
University of Pittsburgh
*milos@cs.pitt.edu*

## Abstract

Approximate linear programming (ALP) is an efficient approach to solving large factored Markov decision processes (MDPs). The main idea of the method is to approximate the optimal value function by a set of basis functions and optimize their weights by linear programming (LP). This paper proposes a new ALP approximation. Comparing to the standard ALP formulation, we decompose the constraint space into a set of low-dimensional spaces. This structure allows for solving the new LP efficiently. In particular, the constraints of the LP can be satisfied in a compact form without an exponential dependence on the treewidth of ALP constraints. We study both practical and theoretical aspects of the proposed approach. Moreover, we demonstrate its scale-up potential on an MDP with more than $2^{100}$ states.

## 1 Introduction

Markov decision processes (MDPs) [19] are an established framework for solving sequential decision problems under uncertainty. Unfortunately, traditional methods for solving MDPs, such as value and policy iteration, are unsuitable for solving real-world problems. These problems are generally structured, and their state and action spaces are represented by state and action variables. The size of these problems is naturally exponential in the number of the variables, and so are their exact solutions. Approximate linear programming (ALP) [21] has emerged as a promising approach to solving these problems efficiently [6, 12, 15].

The main idea of this method is to approximate the optimal value function by a set of basis functions and optimize their weights by linear programming (LP). The optimization can be performed in a structured manner [10, 20]. The structure is a result of combining the structure of factored MDPs and linear value function approximations.

The complexity of computing exact ALP solutions [10, 20] is exponential in the treewidth of the dependency graph that represents the constraint space in ALP. Therefore, when the treewidth of an ALP is large, its exact solution is infeasible. This type of problems can be still solved approximately using Monte Carlo constraint sampling [7, 14]. This approach can be interpreted as an outer approximation to the feasible region of the ALP.

In this work, we propose inner approximations to the feasible region. In comparison to the standard ALP formulation, the constraint space is factored into a set of subspaces. This structure allows for solving the new LP more efficiently. In particular, its constraints can be satisfied in a compact form without an exponential dependence on the treewidth of the original constraint space. We investigate both practical and theoretical aspects of the approach. In addition, we demonstrate that the approach yields an exponential speedup over ALP.

The paper is organized as follows. First, we review factored MDPs [5] and linear value function approximations [2, 22]. Second, we discuss in detail existing work on approximate linear programming. Third, we propose a novel partitioned ALP formulation and study its properties. Finally, we evaluate the quality of the approximation on decision problems with more than $2^{100}$ states.

## 2 Factored MDPs

Many real-world decision problems are naturally described in a factored form. Factored MDPs [5] allow for a compact representation of this structure.

A *factored MDP* [5] is a 4-tuple $\mathcal{M} = (\mathbf{X}, \mathcal{A}, P, R)$, where $\mathbf{X} = \{X_1, \ldots, X_n\}$ is a state space represented by a set of state variables, $\mathcal{A} = \{a_1, \ldots, a_m\}$ is a finite set of actions[1], $P(\mathbf{X}' \mid \mathbf{X}, \mathcal{A})$ is a transition function, which represents the dynamics of the MDP, and $R$ is a reward function assigning immediate payoffs to state-action configurations. The state of the system is completely observed and given by a vector of value assignments $\mathbf{x} = (x_1, \ldots, x_n)$.

---

[1]For simplicity of exposition, we consider an MDP model with a single action variable $\mathcal{A}$. Our ideas straightforwardly generalize to MDPs with factored action spaces [11].

**Transition model:** The transition model is represented by a conditional probability distribution $P(\mathbf{X}' \mid \mathbf{X}, \mathcal{A})$, where $\mathbf{X}$ and $\mathbf{X}'$ denote the state variables at two successive time steps. Since the full tabular representation of $P(\mathbf{X}' \mid \mathbf{X}, \mathcal{A})$ is infeasible when the number of state variables is large, we assume that the distribution factors along $\mathbf{X}'$ as:

$$P(\mathbf{X}' \mid \mathbf{X}, a) = \prod_{i=1}^{n} P(X_i' \mid \mathsf{Par}(X_i'), a) \qquad (1)$$

and is described compactly by a *dynamic Bayesian network (DBN)* [8]. The network reflects independencies among the variables $\mathbf{X}$ and $\mathbf{X}'$ given an action $a$. One-step dynamics of every state variable is given by its conditional probability distribution $P(X_i' \mid \mathsf{Par}(X_i'), a)$, where $\mathsf{Par}(X_i') \subseteq \mathbf{X}$ is the parent set of $X_i'$. The parent set is usually a small subset of state variables which simplifies the parameterization of the model.

**Reward model:** The reward model is factored similarly to the transition model. Specifically, the reward function:

$$R(\mathbf{x}, a) = \sum_{j} R_j(\mathbf{x}_j, a) \qquad (2)$$

is an additive function of local reward functions defined on the subsets $\mathbf{X}_j$ and $\mathcal{A}$. These local functions are compactly represented by reward nodes $R_j$, which are conditioned on their parent sets $\mathsf{Par}(R_j) = \mathbf{X}_j \cup \mathcal{A}$.

**Optimal value function and policy:** The quality of a policy $\pi$ is measured by the *infinite horizon discounted reward* $\mathbb{E}[\sum_{t=0}^{\infty} \gamma^t r_t]$, where $\gamma \in [0, 1)$ is a *discount factor* and $r_t$ is the immediate reward at the time step $t$. In such a setting, there always exists an *optimal policy* $\pi^*$ which is stationary and deterministic [19]. The policy is greedy with respect to the *optimal value function* $V^*$, which is a fixed point of the Bellman equation [1]:

$$V^*(\mathbf{x}) = \max_{a} \left[ R(\mathbf{x}, a) + \gamma \mathbb{E}_{P(\mathbf{x}' \mid \mathbf{x}, a)}[V^*(\mathbf{x}')] \right]. \qquad (3)$$

Similarly to the above equation, all expectation terms in the rest of the paper are written compactly as $\mathbb{E}_{P(\mathbf{x})}[f(\mathbf{x})]$.

## 3 Solving factored MDPs

Markov decision processes can be solved by exact dynamic programming (DP) methods in polynomial time in the size of their state space [19]. Unfortunately, the space space $\mathbf{X}$ of factored MDPs is exponential in the number of state variables. Therefore, the DP methods are unsuitable for solving these problems. Since a factored representation of an MDP does not guarantee a structure in its solution [13], we resort to value function approximations.

In this work, we focus on the *linear value function approximation* [2, 22]:

$$V^{\mathbf{w}}(\mathbf{x}) = \sum_i w_i f_i(\mathbf{x}). \qquad (4)$$

The approximation restricts the form of the value function to the linear combination of basis functions $f_i(\mathbf{x})$, where $\mathbf{w}$ is a vector of optimized weights. The basis functions $f_i(\mathbf{x})$ are arbitrary functions, which are usually restricted to small subsets of state variables $\mathbf{X}_i$ [2, 13]. The functions play the same role as features in machine learning. They are usually provided by domain experts but can also be discovered automatically [18, 16].

## 4 Approximate linear programming

Various techniques for optimizing the linear value function approximation have been studied and analyzed [3]. We focus on *approximate linear programming (ALP)* [21], which restates this problem as a linear program:

$$\begin{aligned}
\text{minimize}_{\mathbf{w}} \quad & \sum_i w_i \alpha_i & (5)\\
\text{subject to:} \quad & \sum_i w_i F_i(\mathbf{x}, a) - R(\mathbf{x}, a) \geq 0\\
& \forall \, \mathbf{x} \in \mathbf{X}, a \in \mathcal{A};
\end{aligned}$$

where $\mathbf{w}$ denotes the variables in the LP, $\alpha_i$ is a *basis function relevance weight*:

$$\alpha_i = \mathbb{E}_{\psi(\mathbf{x})}[f_i(\mathbf{x})], \qquad (6)$$

$\psi(\mathbf{x}) \geq 0$ is a *state relevance density function* that weights the quality of the approximation, and:

$$F_i(\mathbf{x}, a) = f_i(\mathbf{x}) - \gamma \mathbb{E}_{P(\mathbf{x}' \mid \mathbf{x}, a)}[f_i(\mathbf{x}')] \qquad (7)$$

denotes the difference between the basis function $f_i(\mathbf{x})$ and its discounted *backprojection*. This linear program is feasible if the set of basis functions includes a constant function $f_0(\mathbf{x}) \equiv 1$. We assume that such a basis function is present.

Since our basis functions $f_i(\mathbf{x})$ are often restricted to small subsets of state variables, expectation terms in the ALP formulation (5) can be computed efficiently [10]. For instance, the backprojection terms can be rewritten as:

$$\mathbb{E}_{P(\mathbf{x}' \mid \mathbf{x}, a)}[f_i(\mathbf{x}')] = \mathbb{E}_{P(\mathbf{x}_i' \mid \mathbf{x}, a)}[f_i(\mathbf{x}_i')], \qquad (8)$$

where $\mathbf{X}_i'$ is a lower dimensional state space corresponding to the basis function $f_i(\mathbf{x})$, and $P(\mathbf{x}_i' \mid \mathbf{x}, a)$ is a distribution defined on this subspace. Similarly, state relevance weights $\alpha_i$ can be computed efficiently if the state relevance density $\psi(\mathbf{x})$ is structured.

### 4.1 Solving ALP formulations

The major problem in solving ALP formulations efficiently is in satisfying their constraints. This problem is hard since the number of the constraints is exponential in the number of state variables. Fortunately, the constraints exhibit some structure. The structure is a result of combining linear value function approximations (Equation 4) with factored reward and transition models (Equations 1 and 2). Therefore, ALP constraints can be satisfied in a structured form and without being enumerated exhaustively.

Based on these observations, Guestrin *et al.* [10] proposed a variable elimination method [9] that rewrites the constraint space compactly. Schuurmans and Patrascu [20] solved the constraint satisfaction problem by the cutting plane method [4]. The approach iteratively searches for the most violated constraint:

$$\arg\min_{\mathbf{x},a} \left[ \sum_i w_i^{(t)} F_i(\mathbf{x}, a) - R(\mathbf{x}, a) \right] \quad (9)$$

with respect to the solution $\mathbf{w}^{(t)}$ of a relaxed ALP. The most violated constraint is added to the linear program, which is in turn resolved for a new vector $\mathbf{w}^{(t+1)}$. This procedure is iterated until no violated constraint is found. In such a case, the vector $\mathbf{w}^{(t)}$ is an optimal solution to the ALP.

The space complexity of both constraint satisfaction methods [10, 20] is exponential in the treewidth of the constraint space. As a result, the methods are unsuitable for problems with a large treewidth. However, such problems can be still solved approximately. For instance, de Farias and Van Roy [7] proposed Monte Carlo approximations of the constraint space. Kveton and Hauskrecht [14] showed how to search for the most violated constraint (Equation 9) using Markov chain Monte Carlo (MCMC) sampling.

### 4.2 Theoretical analysis

The quality of the ALP formulation has been studied by de Farias and Van Roy [6]. Based on their work, we conclude that ALP minimizes the $\mathcal{L}_1$-norm error $\|V^* - V^{\mathbf{w}}\|_{1,\psi}$. The following theorem draws a parallel between optimizing this objective and the max-norm error $\|V^* - V^{\mathbf{w}}\|_\infty$.

**Theorem 1** (de Farias and Van Roy [6]). *Let $\widetilde{\mathbf{w}}$ be a solution to the ALP formulation (5). Then the expected error of the value function $V^{\widetilde{\mathbf{w}}}$ can be bounded as:*

$$\left\|V^* - V^{\widetilde{\mathbf{w}}}\right\|_{1,\psi} \leq \frac{2}{1-\gamma} \min_{\mathbf{w}} \|V^* - V^{\mathbf{w}}\|_\infty,$$

*where $\|\cdot\|_{1,\psi}$ is an $\mathcal{L}_1$-norm weighted by the state relevance density function $\psi$ and $\|\cdot\|_\infty$ is the max-norm.*

De Farias and Van Roy [6] also proved a tighter version of Theorem 1, which reweights the error $\|V^* - V^{\mathbf{w}}\|_\infty$.

## 5 Partitioned ALP

In this section, we propose a novel approximate linear programming formulation. In comparison to the standard ALP (5), the proposed formulation has an additional structure in its constraint space. The structure allows for controlling the complexity of solving the new LP.

The LP solves a more restrictive problem than the standard ALP. As a result, the formulation can be viewed as an inner approximation to the feasible region of the ALP (Figure 1). This differentiates our work from existing ALP approximations [7, 14]. These approximations are based on constraint sampling. As a result, they approximate the feasible region of the ALP from outside.

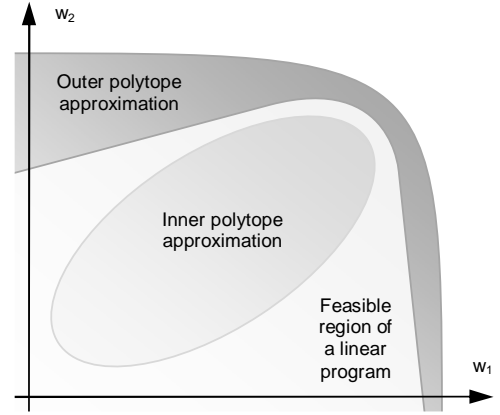

Figure 1: An illustration of inner and outer approximations to the feasible region of a linear program.

### 5.1 An illustrative example

First, let us consider an optimization problem:

$$\begin{aligned}
\text{minimize}_{\mathbf{w},h} \quad & w_1 \alpha_1 + w_2 \alpha_2 + h \quad (10) \\
\text{subject to:} \quad & w_1 F_1(x_1) + w_2 F_2(x_2) + h \geq 0 \\
& \forall\, x_1 \in X_1, x_2 \in X_2;
\end{aligned}$$

where $\mathbf{w} = (w_1, w_2)$ denotes the main optimized variables, and $h$ is an auxiliary variable that guarantees the feasibility of the LP. This problem involves $|X_1 \times X_2| = |X_1| \times |X_2|$ constraints. If the number of the constraints is large, a suboptimal but feasible solution to the problem can be obtained by solving a new linear program:

$$\begin{aligned}
\text{minimize}_{\mathbf{w},h} \quad & w_1 \alpha_1 + w_2 \alpha_2 + h \quad (11) \\
\text{subject to:} \quad & h_1 + h_2 = h \\
& w_1 F_1(x_1) + h_1 \geq 0 \quad \forall\, x_1 \in X_1 \\
& w_2 F_2(x_2) + h_2 \geq 0 \quad \forall\, x_2 \in X_2;
\end{aligned}$$

where $h_1$ and $h_2$ are new auxiliary variables that guarantee the feasibility of the LP. Note that the new LP decomposes the original constraint $w_1 F_1(x_1) + w_2 F_2(x_2) + h \geq 0$ into two smaller constraint spaces with $|X_1| + |X_2|$ constraints. Therefore, it is typically faster to solve the new LP than our original problem (10). In the next section, we show how to apply similar ideas in the context of ALP.

### 5.2 Partitioned ALP formulation

Similarly to Section 5.1, we may decompose the constraint space in the ALP formulation (5). Formally, the *partitioned*

ALP (PALP) formulation with $K$ constraint spaces is given by a linear program:

$$\text{minimize}_\mathbf{w} \quad \sum_i w_i \alpha_i \qquad (12)$$
$$\text{subject to:} \quad \mathbf{D}\mathbf{M}_\mathbf{w}(\mathbf{x}, a)^\top \geq 0 \quad \forall\, \mathbf{x} \in \mathbf{X}, a \in \mathcal{A};$$

where:

$$\mathbf{M}_\mathbf{w}(\mathbf{x}, a) = (w_1 F_1(\mathbf{x}, a), \ldots, -R_1(\mathbf{x}_1, a), \ldots) \qquad (13)$$

is a vector whose $i$-th element corresponds to the $i$-th term in the ALP constraint, and the *partitioning matrix*:

$$\mathbf{D} = \begin{pmatrix} d_{1,1} & d_{1,2} & d_{1,3} & \cdots \\ d_{2,1} & d_{2,2} & d_{2,3} & \cdots \\ d_{3,1} & d_{3,2} & d_{3,3} & \cdots \\ \vdots & \vdots & \vdots & \ddots \end{pmatrix} \qquad (14)$$

determines how the ALP constraint decomposes into the $K$ new constraint spaces. Specifically, the term $d_{k,i}$ measures the contribution of the $i$-th term in the ALP constraint to the $k$-th constraint space. Due to this interpretation, we assume that all terms $d_{k,i}$ are non-negative and that the partitioning matrix $\mathbf{D}$ is normalized such that the equality $\sum_k d_{k,i} = 1$ holds for all $i$. Under such assumptions, it is trivial to show that the satisfaction of the $K$ constraints $\mathbf{D}\mathbf{M}_\mathbf{w}(\mathbf{x}, a)^\top \geq 0$ leads to the satisfaction of a corresponding ALP constraint. The claim can be proved based on the identity:

$$\mathbf{1}\mathbf{D}\mathbf{M}_\mathbf{w}(\mathbf{x}, a)^\top = \sum_i w_i F_i(\mathbf{x}, a) - R(\mathbf{x}, a), \qquad (15)$$

where $\mathbf{1}$ is a row vector of ones. It follows that every PALP solution is feasible in a corresponding ALP.

Similarly to ALP, the feasibility of the PALP formulation is guaranteed if the set of basis functions includes a constant function $f_0(\mathbf{x}) \equiv 1$. We assume that the function is present in all $K$ constraint spaces. In each of them, we define a new weight $w_0^k$, which reflects the contribution of this function. As a result of these changes, the PALP formulation slightly changes its form:

$$\text{minimize}_\mathbf{w} \quad \sum_i w_i \alpha_i + w_0 \qquad (16)$$
$$\text{subject to:} \quad \sum_k w_0^k = w_0$$
$$\mathbf{D}\mathbf{M}_\mathbf{w}(\mathbf{x}, a)^\top + (1-\gamma)(w_0^1, \ldots, w_0^K)^\top \geq 0$$
$$\forall\, \mathbf{x} \in \mathbf{X}, a \in \mathcal{A}.$$

In the rest of the paper, we use the above and original PALP formulations interchangeably.

### 5.3 Partitioning matrix

The partitioning matrix $\mathbf{D}$ allows for trading off the quality and complexity of PALP solutions. To achieve high-quality and tractable approximations, the rows of the matrix should reflect tree decompositions of the *cost network* corresponding to ALP constraints (Figure 2). The width of the decompositions should be small since the complexity of satisfying a single constraint space is exponential in its treewidth [10].

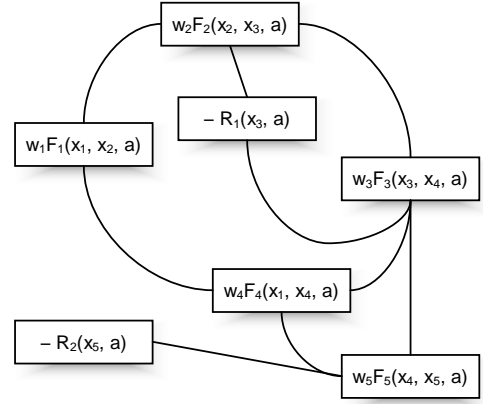

Figure 2: A graphical representation of a cost network. The rectangular nodes represent functions, which are defined on some subset of variables. Two nodes in the cost network are connected if their functions share at least one variable.

How to generate the best PALP approximation within a certain complexity limit is an open question. In the experimental section, we build the matrix $\mathbf{D}$ based on a heuristic. The heuristic generates a constraint space for every expectation term $F_k(\mathbf{x}, a)$ in Equation 9. This constraint space consists of the term $w_k F_k(\mathbf{x}, a)$ and its cost network neighbors. The constraint space is not included in the matrix $\mathbf{D}$ if its terms constitute a subset of another constraint space.

This decomposition of our initial problem can be viewed as optimizing $K$ smaller MDPs, which have overlapping state and action spaces, and share value functions. To clarify the construction of the matrix $\mathbf{D}$, we demonstrate it on the cost network in Figure 2. The cost network involves 7 functions, out of which 5 have the form of $w_k F_k(\mathbf{x}, a)$. Therefore, the corresponding matrix $\mathbf{D}$ has 5 rows and 7 columns:

$$\mathbf{D} = \begin{pmatrix} 0.3\bar{3} & 0.3\bar{3} & 0 & 0.25 & 0 & 0 & 0 \\ 0.3\bar{3} & 0.3\bar{3} & 0.25 & 0 & 0 & 0.5 & 0 \\ 0 & 0.3\bar{3} & 0.25 & 0.25 & 0.3\bar{3} & 0.5 & 0 \\ 0.3\bar{3} & 0 & 0.25 & 0.25 & 0.3\bar{3} & 0 & 0 \\ 0 & 0 & 0.25 & 0.25 & 0.3\bar{3} & 0 & 1 \end{pmatrix}. \qquad (17)$$

Non-zero entries $d_{k,i}$ in the matrix indicate that the $i$-th cost network term is present in the $k$-th constraint space.

### 5.4 Solving PALP formulations

The PALP formulation (12) is similar to the ALP formulation (5). As a result, it can be solved in a similar fashion. In the experimental section, we implemented the cutting plane method for solving linear programs (Figure 3). In principle, any method for solving ALPs (Section 4.1) can be adapted to PALPs.

**Inputs:**
    a factored MDP $\mathcal{M} = (\mathbf{X}, \mathcal{A}, P, R)$
    basis functions $f_0(\mathbf{x}), f_1(\mathbf{x}), f_2(\mathbf{x}), \ldots$
    initial basis function weights $\mathbf{w}^{(0)}$
    a separation oracle $\mathcal{O}$

**Algorithm:**
    initialize a relaxed PALP formulation
    $t = 0$
    while a stopping criterion is not met
      for every constraint space $k = 1, \ldots, K$
        query the oracle $\mathcal{O}$ for a violated constraint $(\mathbf{x}_\mathcal{O}, a_\mathcal{O})$
        if the constraint $(\mathbf{x}_\mathcal{O}, a_\mathcal{O})$ is violated
          add the constraint to the relaxed PALP
        resolve the LP for a new vector $\mathbf{w}^{(t+1)}$
      $t = t + 1$

**Outputs:**
    basis function weights $\mathbf{w}^{(t)}$

Figure 3: Pseudo-code implementation of the cutting plane method for solving PALP formulations.

### 5.5 Theoretical analysis

In this section, we discuss the quality of the PALP formulation (12). First, we prove that its solution is an upper bound on the optimal value function $V^*$.

**Proposition 1.** *Let $\widetilde{\mathbf{w}}$ be a solution to the PALP formulation (12). Then $V^{\widetilde{\mathbf{w}}} \geq V^*$.*

**Proof:** Since $\widetilde{\mathbf{w}}$ is a solution to the PALP formulation (12), it is also a suboptimal solution to the ALP formulation (5). Therefore, the constraint $V^{\widetilde{\mathbf{w}}} \geq \mathcal{T}^* V^{\widetilde{\mathbf{w}}}$ is satisfied. Furthermore, note that the Bellman operator $\mathcal{T}^*$ is both monotonic and contracting. Hence, the inequality $V^{\widetilde{\mathbf{w}}} \geq \mathcal{T}^* V^{\widetilde{\mathbf{w}}}$ yields the following sequence of inequalities:

$$V^{\widetilde{\mathbf{w}}} \geq \mathcal{T}^* V^{\widetilde{\mathbf{w}}} \geq \mathcal{T}^* \mathcal{T}^* V^{\widetilde{\mathbf{w}}} \geq \cdots \geq V^*.$$

This step concludes our proof. ∎

The above result allows us to restate the objective $\mathbb{E}_\psi[V^\mathbf{w}]$ in PALP.

**Proposition 2.** *The objective in the PALP formulation (12) can be rewritten as $\|V^* - V^\mathbf{w}\|_{1,\psi}$, where $\|\cdot\|_{1,\psi}$ is an $\mathcal{L}_1$-norm weighted by the state relevance density function $\psi$.*

**Proof:** Follows from the fact that all solutions to the PALP formulation (12) satisfy the constraint $V^\mathbf{w} \geq V^*$. ∎

Based on Proposition 2, we conclude that PALP optimizes the linear value function approximation with respect to the reweighted $\mathcal{L}_1$-norm error $\|V^* - V^\mathbf{w}\|_{1,\psi}$. The following theorem draws a parallel between optimizing this objective and the max-norm error $\|V^* - V^\mathbf{w}\|_\infty$.

**Theorem 2.** *Let $\widetilde{\mathbf{w}}$ be a solution to the PALP formulation (12). Then the expected error of the value function $V^{\widetilde{\mathbf{w}}}$ can be bounded as:*

$$\left\|V^* - V^{\widetilde{\mathbf{w}}}\right\|_{1,\psi} \leq \frac{2}{1-\gamma} \min_\mathbf{w} \|V^* - V^\mathbf{w}\|_\infty + \frac{K\delta}{1-\gamma},$$

*where $\|\cdot\|_{1,\psi}$ is an $\mathcal{L}_1$-norm weighted by the state relevance density function $\psi$, $\|\cdot\|_\infty$ is the max-norm, $\delta$ is a scalar that reflects how hard is to make an ALP solution feasible in our PALP formulation, and $K$ denotes the number of constraint spaces in the PALP.*

**Proof:** Our proof is similar to the proof of Theorem 2 by de Farias and Van Roy [6]. The vectors $\widetilde{\mathbf{w}}$, $\overline{\mathbf{w}}$, and $\mathbf{w}^*$ denote an optimal solution to the PALP formulation, its suboptimal solution, and the vector that minimizes the max-norm error $\|V^* - V^\mathbf{w}\|_\infty$, respectively. First, we bound the objective in the PALP as follows:

$$\left\|V^* - V^{\widetilde{\mathbf{w}}}\right\|_{1,\psi} \leq \left\|V^* - V^{\overline{\mathbf{w}}}\right\|_{1,\psi}$$
$$\leq \left\|V^* - V^{\overline{\mathbf{w}}}\right\|_\infty.$$

Second, we bound the max-norm error of $V^{\overline{\mathbf{w}}}$ by the triangle inequality:

$$\left\|V^* - V^{\overline{\mathbf{w}}}\right\|_\infty \leq \left\|V^* - V^{\widehat{\mathbf{w}}}\right\|_\infty + \left\|V^{\widehat{\mathbf{w}}} - V^{\overline{\mathbf{w}}}\right\|_\infty,$$

where $\widehat{\mathbf{w}}$ is an arbitrary solution to the ALP formulation. In the rest of the proof, we bound the two terms on the right-hand side of the inequality. The first term reflects how hard is to fit the linear value function approximation to the value function $V^*$. If the vector $\widehat{\mathbf{w}}$ is set such that:

$$\widehat{\mathbf{w}} = \mathbf{w}^* + \frac{1+\gamma}{1-\gamma} \left\|V^* - V^{\mathbf{w}^*}\right\|_\infty i_0,$$

where $i_0 = (1, 0, \ldots, 0)$ is an indicator of the constant basis function $f_0(\mathbf{x}) \equiv 1$, the following inequality:

$$\left\|V^* - V^{\widehat{\mathbf{w}}}\right\|_\infty \leq \frac{2}{1-\gamma} \left\|V^* - V^{\mathbf{w}^*}\right\|_\infty$$

holds [6]. The second term reflects how hard it to make the ALP solution $\widehat{\mathbf{w}}$ feasible in the PALP. If the vector $\overline{\mathbf{w}}$ is set such that:

$$\overline{\mathbf{w}} = \widehat{\mathbf{w}} + \frac{K\delta}{1-\gamma} i_0,$$

where $i_0 = (1, 0, \ldots, 0)$ is an indicator of the constant basis function $f_0(\mathbf{x}) \equiv 1$, $\delta = -\min_{\mathbf{x},a} \min(\mathbf{DM}_{\widehat{\mathbf{w}}}(\mathbf{x}, a)^\intercal)$, and the function $\min(\mathbf{DM}_{\widehat{\mathbf{w}}}(\mathbf{x}, a)^\intercal)$ computes the minimum of the vector $\mathbf{DM}_{\widehat{\mathbf{w}}}(\mathbf{x}, a)^\intercal$, we can guarantee the feasibility of $\overline{\mathbf{w}}$. The proof is based on the observation that all constraints in the feasible PALP formulation (16) are satisfied when the weights $\overline{w}_0^k$ are set such that:

$$\overline{w}_0^k = \frac{1}{K} \widehat{w}_0 + \frac{\delta}{1-\gamma}.$$

Based on this setting, the max-norm error between $V^{\widehat{\mathbf{w}}}$ and $V^{\overline{\mathbf{w}}}$ is bounded as:

$$\left\|V^{\widehat{\mathbf{w}}} - V^{\overline{\mathbf{w}}}\right\|_\infty \leq \frac{K\delta}{1-\gamma}.$$

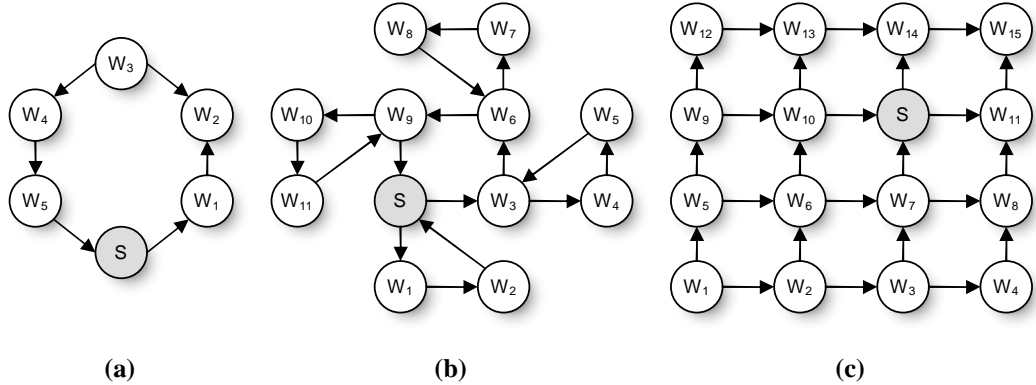

Figure 4: An illustration of three network administration topologies: **a.** 6-ring, **b.** 12-ring-of-rings, and **c.** $4 \times 4$ grid. The gray and white nodes represent the server and workstations, respectively. The computers are connected along the arrows.

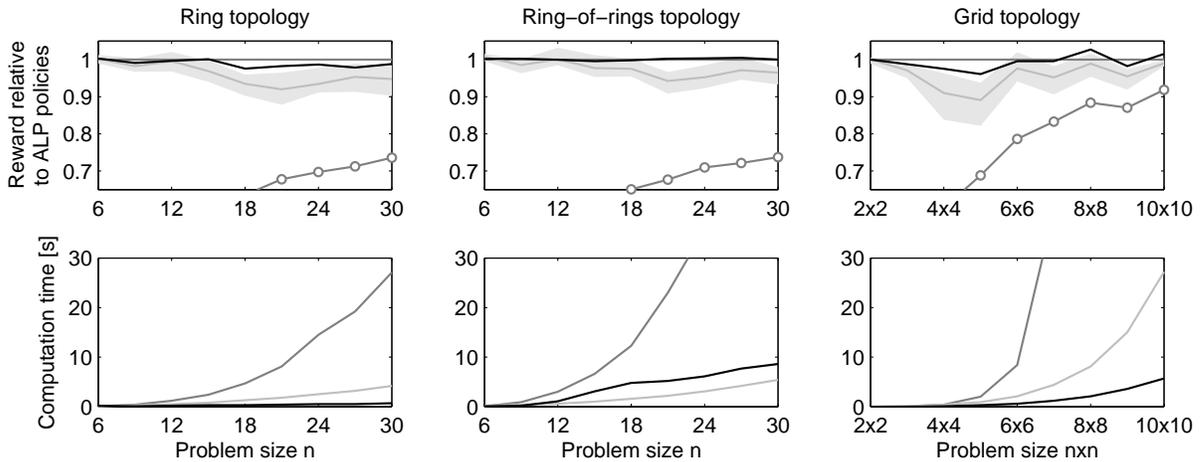

Figure 5: Comparison of four policies for solving the network administration problem. The first policy is obtained by PALP (black lines), the second one by ALP (dark gray lines), the third one by ALP with randomly sampled constraints (light gray lines), and the fourth policy is the server heuristic (dark gray lines with circles). The policies are compared by their reward, which is measured relatively to the reward of ALP policies, and computation time (in seconds). The variance in the rewards of sampled ALP approximations is depicted by gray areas. All results are reported as functions of increasing problem sizes ($n$).

This step concludes our proof. ∎

The above result can be interpreted as follows. PALP yields a close approximation $V^{\widetilde{\mathbf{w}}}$ to the optimal value function $V^*$ if the function $V^*$ lies in the span of basis functions and the penalty $\delta$ for partitioning the ALP constraint space is small. Unfortunately, we do not have a good bound for the penalty term $\delta$. The value of $\delta$ can be as bad as $\|\widehat{\mathbf{w}}\|_1 + R_{\max}$, where $R_{\max}$ denotes the maximum immediate reward in an MDP. Hence, the bound in Theorem 2 is not very tight in practice. Nevertheless, it provides valuable insights into two sources of errors for PALP approximations.

## 6 Experiments

The objective of the experimental section is to demonstrate the quality and scale-up potential of PALP approximations. The approximations are studied with respect to ALP, which is a state-of-the-art approach to solving large-scale factored MDPs. Our experiments are performed on various forms of the network administration problem [10]. This is a standard benchmark for testing the scalability of MDP algorithms.

### 6.1 Experimental setup

The network administration problem involves a network of randomly crashing computers. When a computer crashes, it increases the probability of its network neighbors crashing. The objective is to reboot crashed computers to restore their functionality and prevent further spreading of their failures into the network. Examples of three network topologies are shown in Figure 4. Each network consists of one server and several workstations. The difference between the two types of the computers is in the reward for keeping them running.

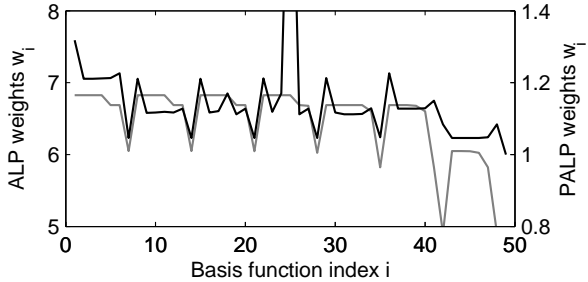

Figure 6: Basis function weights $w_i$ obtained by ALP (dark gray line) and PALP (black line) on the $7 \times 7$ grid network administration problem.

The immediate reward for keeping a workstation running is 1. The reward for keeping the server running is 2.

The network administration problem is a challenging MDP due to the size of its state space. Specifically, since the state of the network is a product of individual computer states, it is exponential in the number of computers. Therefore, only small instances of the problem can be solved exactly. In the rest of the section, we focus on large-scale problems and try to solve them through linear value function approximations (Equation 4). In all experiments, we define a basis function $f_i(\mathbf{x}) = x_i$ for every computer $X_i$. Furthermore, in the ring and ring-of-rings topologies (Figures 4a and 4b), we assign a pairwise basis function $f_{i \to j}(\mathbf{x}) = x_i x_j$ to every network connection $X_i \to X_j$.

Our linear value function approximations are optimized using ALP and PALP formulations. The cutting plane method is employed to solve these LPs exactly and efficiently (Figure 3). In addition, we experiment with ALP formulations, which are solved approximately by Monte Carlo constraint sampling [7]. The number of sampled constraints is $100n$, where $n$ is the number of state variables $\mathbf{X}$. Therefore, it is proportional to the size of solved problems. To demonstrate the non-triviality of learned policies, we also report results of a heuristic for solving our problem. The heuristic places the administrator at the server so the computer is protected from crashing.

### 6.2 Experimental results

Our main experimental results are summarized in Figure 5. Based on these results, we conclude that PALP policies are almost as good as ALP policies. Specifically, note that the rewards of the policies are within 95 percent of our baseline in all experiments. Unfortunately, these good results cannot be explained by Theorem 2 because our bound is too loose. To explain our results, we tried to investigate the similarity of basis function weights $\mathbf{w}$ obtained by ALP and PALP. As illustrated in Figure 6, the magnitudes of the weights can be very different. However, the weights exhibit similar trends. In turn, value function approximations corresponding to the weights must have similar shapes, and their greedy policies

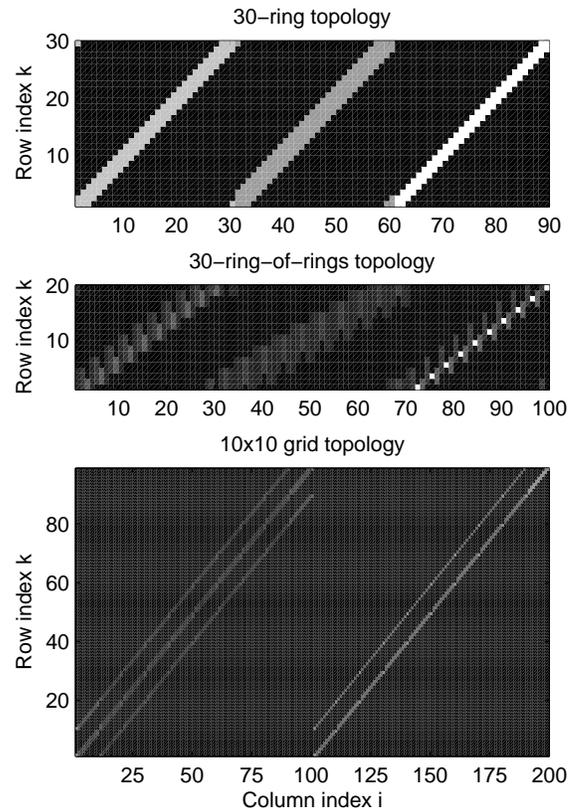

Figure 7: Three partitioning matrices $\mathbf{D}$ corresponding to the network administration problem. The brighter the color of a pixel, the higher the value of the partitioning coefficient $d_{k,i}$. Black pixels represent zero coefficients.

are similar as a result.

Figure 5 also suggests that PALP policies can be computed significantly faster than ALP policies. This speedup results from working with sparse decompositions (Figure 7) of the original constraint space rather than the space itself. Moreover, note that the treewidth of the $n \times n$ network administration problem (Figure 4c) is $n$. Therefore, the complexity of learning ALP policies for this problem is naturally exponential in $n$. On the other hand, the complexity of learning PALP policies is polynomial in $n$. This claim follows from the observation that the number of PALP constraint spaces is $n^2$ and their treewidth is not dependent on $n$. As a result, PALP on the grid network provides an exponential speedup over ALP. This result can be verified by the analysis of the computation time trends in Figure 5.

Finally, Figure 5 illustrates that PALP policies are superior to ALP policies, which are obtained by ALP with randomly sampled constraints. In most cases, the PALP policies yield significantly higher rewards than the average sampled ALP approximation. For all larger network administration problems, the policies are as good or better than the best of these approximations. At the same time, the computation time of the PALP policies is shorter or comparable to the computation time of the sampled approximations.

## 7 Conclusions

Development of scalable algorithms for solving real-world MDPs is a challenging task. In this work, we investigated a novel approach to approximate linear programming. Comparing to the standard ALP formulation, we decompose the constraint space into a set of low-dimensional spaces. This structure allows for solving the new LP more efficiently. In particular, its constraints can be satisfied in a compact form without an exponential dependence on the treewidth of the original constraint space. Our experiments demonstrate the superiority of the new approach when compared to existing exact and approximate solutions to ALP.

Results of this paper can be extended in several ways. First, we have not addressed the topic of learning good partitioning matrices $\mathbf{D}$. This topic is in many aspects similar to the problem of efficient inference in Bayesian networks. In this context, Meila [17] proposed using a mixture of trees to approximate an arbitrary joint probability distribution defined by a Bayesian network. Second, the bound in Theorem 2 is definitely loose in practice. How to make this bound tight is an interesting open question. Finally, PALP and its benefits should be studied on a more realistic problem than the one presented in the experimental section.

## Acknowledgment

We thank anonymous reviewers for helpful comments that led to the improvement of this paper. We also thank Carlos Guestrin for encouragement and positioning this paper in a broader context.